# AGI and reflexivity

## Pascal Faudemay[1]

**Abstract.** We define a property of intelligent systems, which we call Reflexivity. In human beings it is one aspect of consciousness, and an element of deliberation.

We propose a conjecture, that this property is conditioned by a topological property of the processes which implement this reflexivity. These processes may be symbolic, or non symbolic e.g. connexionnist. An architecture which implements reflexivity may be based on the interaction of one or several modules of deep learning, which may be specialized or not, and interconnected in a relevant way. A necessary condition of reflexivity is the existence of recurrence in its processes, we will examine in which cases this condition may be sufficient.

We will then examine how this topology will make possible the expression of a second property, the deliberation. In a final paragraph, we propose an evaluation of intelligent systems, based on the fulfillment of all or some of these properties.

## 1. INTRODUCTION

### 1.1. Focusing on high-level functions

The goal of Artificial General Intelligence (AGI) is to model human intelligence (or at first time intelligence of superior animals), and to make available in a computer system the capabilities of a human being. AGI differ in this way from Specialized Artificial Intelligence, which targets more limited application domains [6, 11, 13].

An AGI may also designate a prototype of an Artificial General Intelligence, more or less close to the goals of this field [4, 12, 35]. It may also be a seed AGI (or Baby AGI), with a limited initial knowledge but a good learning capability [30, 31, 40].

Among the functions of human mind, we consider "superior" functions, including goals management, attention, emotions, inference and analogy, conceptual blending, deliberation. Some of them are mainly unconscious, such as attention and part of inference. Others are part of consciousness, like deliberation.

## 1.2. An axiomatic approach

We propose an axiomatic approach, and therefore it should be as simple as possible. However, let's notice that we do not consider the topic of consciousness as a whole, but only an aspect, which we shall call reflexivity.

**Definition 1.**
*The reflexivity of a cognitive process will be defined as the fact for this process, to be a function of the previous states of the same process, and of its present and previous perceptions, actions and emotions.*

If the previous states which are considered begin at the origin of life of the system, we shall say that this reflexivity is unlimited, and if these states begin at the origin of the considered data system, we shall say that this reflexivity is infinite.

The reference to emotions is explained at paragraph 4.1.

**Definition 2.**
*Deliberation is the production of a flow of sentences in a natural or artificial language (or of tuples which represent sentences, or of lists of concepts), by a reflexive system in which these sentences are part of the future cognitive processes.*

We first present in a more systematic way reflexivity and deliberation, then we show how the implementation of reflexivity implies recurrence. We then discuss which modalities of recurrence guarantee reflexivity, and at which conditions. The recurrence phenomenon may be observed with a partially or entirely non symbolic architecture, but also with an architecture based on the management of symbolic data (sentences or tuples). We also propose some aspects of the recurrent management of deliberation. A last paragraph concludes on the limits of reflexivity as a function of the complexity of the underlying architecture, and proposes an evaluation scheme for reflexive and deliberative AGIs.

## 2. REFLEXIVITY.

An intelligent system implemented by a neural network [19] is reflexive if it has the following properties:

$$S_t = H(W_{t-1}, P, E, A, S_{t-1}) \qquad (1)$$
$$K_t = F(\,Wo, P, E, A, S). \qquad (2)$$
$$W_t = G(W_{t-1}, P, E, A, S). \qquad (3)$$

And the following one:

$$Sj_{=0,\,t+1} = T(S_{j=jmax,\,t}, P, E, A). \quad (4)$$

Contact: pascal.faudemay @ gmail.com

---

[1] Previously at LIP6, Paris, France. Presently retired.

W, P, E, A, S, K, are matrices :

P (perceptions)
W (synaptic weights)
A (bias or activation thresholds)
S (states of neurons)
K (actions of the system)
E (emotions)

These matrices have the following dimensions:
- Time t=0 to i
- Layers j=0 to jmax
- Neurons k=0 to kmax(j) in layer j
- For weights w, the index of targets neurons k'=0 to k'max(j+1)

We shall note Wo as a weights matrix at time t=0, j the index of a neurons layer, jmax the index of the upmost neurons layer (which is projected by recurrence on layer j=0), k the index of a neuron in layer j.

When j=0 then $S_j$, = P and when j=p then K= $S_j$
Therefore (2) is directly derived from (1). (3) does not seem to be a direct condition of reflexivity.

Equation (1) and (4) define a topology of the neural network, or of a symbolic system with comparable properties.

*Here by topology we mean a set of properties which can be expressed as properties of a related graph.*

The properties displayed by equations (1) to (4) define the network as a recurrent neural network (RNN) [19, 21]. However, these conditions (or equivalent ones) are probably not sufficient.

```
<page>
        <title>Antichrist</title>
        <id>865</id>
        <revision>
          <id>15900676</id>
          <timestamp>2002-08-
      03T18:14:12Z</timestamp>
          <contributor>
            <username>Paris</username>
            <id>23</id>
          </contributor>
          <minor />
          <comment>Automated
      conversion</comment>
            <text
      xml:space="preserve">#REDIRECT
      [[Christianity]]</text>
          </revision>
</page>
```

**Fig. 1.** XML produced by a RNN. (Andrej Karpathy, 2015, [19]).

Our project is to reach the definition of sufficient conditions for the implementation of reflexivity in a neuron network. This does not forbid its implementation with a symbolic architecture, possibly with an equivalent topology.

With suitably chosen parameters, and a relevant network topology and size, it is possible to teach a neuron network how to invent XML structures, random but realistic Wikipedia pages, etc. Figure 3 displays an example proposed by Karpathy, of an XML file produced by such a system.

This example shows the capability of the deep RNN (i.e. a RNN with many layers, typically 20 to 150 [21]) to learn both the grammar of an XML structure, and the semantic or the vocabulary of a specific field. Let's remember that data are input character by character, and that the system has no prior knowledge of XML or the utilized language.

## 3. DELIBERATION

Deliberation is the fact for the system to infer sentences based on the ones which it has already produced by reflexive processes, to produce such sentences and to decide which sentences among them, it can output towards the external world.

Thus it is a level of meta-inference, which can be implemented by a deep learning system, or by a symbolic process, e.g. derived from one of the existing pre-AGI projects [6, 13].

We will assume that the deliberation is done by a neuronal subsystem, referenced as B, which receives as input the productions of the previously defined reflexive system, named A. The system will produces as an output a flow of sentences (or more generally files) or actions, by which it communicates with its environment.

In Figure 2 (below) we have :

IN the input unit, of one character (one byte, or possibly two bytes if we want to allow complex Unicode characters)

$a_0....a_n$: a 1D systolic vector
A and B: two recurrent neuron networks, where the output layer is projected on the input layer.
$K_1$ to $K_n$ output units, with actions $L_1$ to $L_j$, and the sentence $L_{j+1}$ to $L_n$. This device (or part of it) can possibly be a 1D systolic vector.
OUTPUT output character of network B, and/ or a Boolean which can control the output character from device A.

We do not represent the register stack which may manage a short term memory for unit A input or output, and possibly also two other register stacks for unit B, under the control of dedicated actions from these two units. The access of a neural network to external memories is subject to active research, which we briefly mention in paragraph 4.14. These external memories do not necessarily need to be limited to a register stack [18, 20, 42].

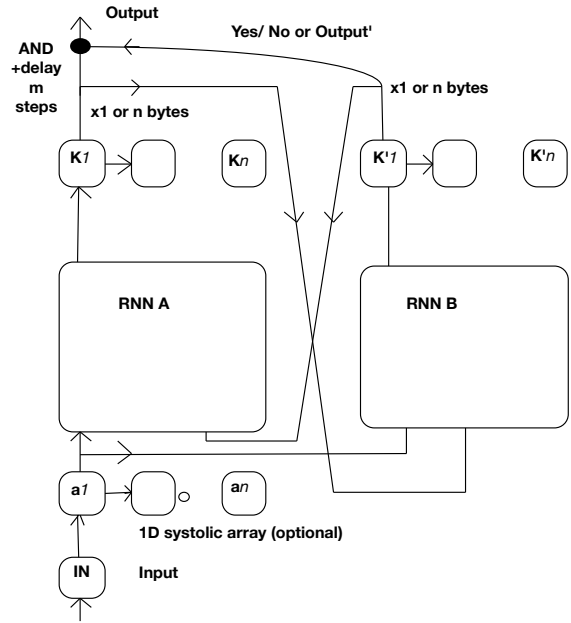

**Fig. 2.** Communication between subsystems A and B.

In order for process B to be effective, it must have the same knowledge of the outside world as the reflexive system A, and be itself a reflexive system. It must also have one or more supplementary neuron layers, connected to a single decision neuron (or to a group of neurons with majority decision), which will produce the result <OUTPUT> or <DON'T OUTPUT>. It may also directly produce one or several sentences of its internal dialogue.

*For this purpose, one solution is that system B would have an initial configuration equivalent to that of system A, and would learn from the outside world in an identical way as A. We shall leave to further research to know whether B must, or must not, be identical to A.*

Let us notice that the B unit processing time implies a minimum delay of at least m steps between output K by unit A, and the same output by the system. We will have

m = n + jmax' (5)

where n is the number of input bytes needed by unit B to decide, and jmax' is the number of layers of unit B. As B is a recurrent network, the effective number of steps for its decision is variable. A step duration is the number of instructions (or of cycles) needed by a layer to process its inputs.

## 4. MODALITIES

### 4.1. Emotions.

In human beings as well as in animals, emotions are conscious states which are accompanied by a sensation, notably of pleasure or displeasure, and which have an effect on memorization. Emotions have a degree or an intensity. Perceptions and states of consciousness which are accompanied by a more intense emotion, are more memorized than those accompanied by a less intense emotion [9, 24, 26].

These consciousness states are the result or the cause of the presence in brain of neurotransmitters or neuromediators. Lövheim [22] proposes a representation of emotions by a cube, as a function of the presence or absence of three neuromediators, dopamin, noradrenalin and serotonin, which have a specially important role. However, there would be a thousand chemical substances at the synaptic junction, and at least one hundred of them would be neurotransmitters or neuromediators [37].

As an example, we can add to those which are cited by Lövheim, endorphins, cannabinoïds and ocytocin. Endorphins modulate pain and well-being, cannabinoïds act on serenity and the sensation of paranoïa (belief in others hostility), ocytocin regulates attachment and indifference.

We speak about human or animal emotions. Indeed, the neuromediators which are present in humans are also present in mammals, and fear or even anxiety can also be observed, even in drosophila [10].

It is important to simulate emotions in an intelligent system, first as a modulator of memorization. A human being without emotions is also at risk of becoming a psychopath. We cannot exclude that a similar phenomenon might appear in artificial intelligence. As human or animals emotions are a powerful modulator of decisions, their absence may lead to inappropriate ones. Last, the presence of emotions in an intelligent system may facilitate its communication with human beings.

This does not necessarily imply that all human or animal emotions should be found in a robot or in an intelligent system. Deciding which are the relevant or minimal emotions for such a system, is outside the scope of this paper.

In an intelligent system an emotion $e_{te}$ with index e (which characterizes the type of emotion) at time t, can be seen as an environment variable defined as the sum of the outputs of neurons of index e, of layers 1 to j, or in a simpler way as the state of neuron of index e in layer jmax. It will be used as the bias of a collection of neurons from various layers.

Emotion $E_e$ activates a collection of $E_e$-dependent neurons, which have a non-null input weight for their connection with neuron of index e. A neuron may be dependent on several emotions, which can either lower or increase its bias, and therefore have an action of inhibition or activation of this neuron.

The system of emotions can also include some coupling between different emotions, some of them may reinforce or inhibit other ones. This is implemented by the production of emotions by a subset of the neuronal network. Any neuron can contribute to the production of some emotion (or of its mediators), and can also produce several emotions. Last, it is possible to introduce some inertia in the emotions, so that it would not change too fast.

It is also possible to associate each emotion with a sense of "pleasure", which will activate a process of continuation of this emotion, or a sense of "displeasure", which will activate a process to end this emotion. These processes may be a simple positive or negative reinforcement on appropriate layers, or a predefined procedural resource, possibly modified by an evolutionary program.

These emotions may also be input into the system by an outside agent (the designer or an user), or be deduced from the activation of some concepts in the neural network.

## 4.2. Cognitive areas, action on bias.

Several subsets of neurons are characterized by their bias, i.e. by an activation threshold which is a function of the level of one or several emotions [26]. In each of these subsets, which we call a cognitive area, the bias is equal for all the neurons of the subset, or at least, it depends on a same function of the emotions.

E.g. in a cognitive area associated with the attachment towards a person or a human group, all neurons will have a bias which will be a function of this attachment. The other properties of cognitive areas, will be discussed shortly in the paragraph about the system initialization.

## 4.3. Research and approximate inference, blending

This property is implicit in a system based on a neuronal architecture, which can easily implement associative retrieval and approximate inference. However, our proposal may also apply to a symbolic architecture.

For examples of approximate inference and of blending in symbolic or partially symbolic architectures, the reader may refer to the works of projects like NARS or OpenCog [12, 36].

The interest of a neuronal approach for AGI is underlined by the works of Schmidhuber [32].

## 4.4. Goals management

We introduce here the notion of an autonomous system.

**Definition 3.**
*An autonomous reflexive system is a reflexive system with a set of actions, which depend on a system of goals.*

We shall notice that the property of deliberation may not be mandatory in a non-reflexive autonomous system, e.g. In the case of an autonomous car. However, we only consider here reflexive and deliberative systems.

The definition of a goals system for an AGI is a complex task. According to Yudkowsky and others, it is very difficult to define them in a non ambiguous way. There would often be a risk that the system would escape the control of its designer, as people do not necessarily understand the reasoning which can be made by an intelligent system, either when it is implemented with a neuronal architecture, or even with a symbolic one [5, 7, 15, 38, 39].

In a very interesting approach, Joscha Bach [3, 4] proposes to derive at any time the goals of an AGI from a set of needs, which include the needs of existence (access to calculation means, energy, cooling,…), social needs (communication with third parties, either human or not,..) and epistemic needs (acquisition of new knowledge). The insufficient satisfaction of a need creates a goal, which can be more or less prioritary.

The MicroPsi system created by Bach is based on the Psi theory proposed by Dietrich Dörner [3]. This theory may enable AGI designers to escape to the various problems of a direct goals definition [39].

## 4.5. Attention

Attention [16] is a property which enables the system to focus its perceptions on some part of the perception field, or to focus inference on part of the sentences or concepts produced by the system. E.g., if the system has a vision capability, with some part in high definition, and some part in low definition, the informations of the low definition part will enable the system to focus the high definition part within the vision field [27].

An example of this property is given by Andrej Karpathy [19], for the access to hand-written characters which are spread on a page, by using a convolutive neural network with a size which is independent of the size of the page.

Attention may also exist for a system which operates on texts, e.g. belonging to a texts database. It may then make actions such as right or left move in the text, text read, move from one document to the another one, etc. The text database can also be more generally considered as a tree, with operators to move within this tree.

This property will then be represented by a set of actions, which will be learned by the neural network in the same way as other actions. A small number of attention properties may also be implemented as "innate properties", e.g, in a procedural way, before starting the system

## 4.6. Perceptions estimation

We define :
$P_{t,1..p}$ the p elementary perceptions at time t, they may be pixels (r, g, b) in a matrix representing an image, or characters in a text, or phonemes in an audio flux, etc.

Let then be $P_{t+2jmax, \ p+1..2p}$ the estimations of these perceptions at time t+2jmax, where jmax is the number of layers both in encoder and in decoder part of the network.

In each recurrent unit A and B, there are two blocks of layers, one of them acts as an encoder which outputs a series of possible actions K, the next ones acts as a decoder which estimate the next state of perceptions. The difference between estimated and real perceptions can produce an increase in attention, or various emotions (surprise, interest, etc..). These effects are now well known for human beings.

In a simplified version of the system layers p+jmax+1 to p+2jmax do not exist, $P_{t+2jmax}$ is replaced by a subset of $S_{jmax}$.

We assume that the system learns by gradient retro-propagation or any other learning method, which is done from jmax to 1 for the encoder part, and from 2jmax to jmax+1 for the decoder part, if it exists. For this part, the $W_t$ weights are modified as a function of the difference between the predicted and observed perceptions.

## 4.7. Models of others and of itself

We assume that the system encodes an internal discourse, composed of trees of sentences, as trees of tuples, where a sentence can be represented by several tuples. In these tuples, either by construction or by learning, there is a field or attribute which designates the author of the sentence (or the assumption which it represents) and attributes of place and time or duration.

An author designation may be done explicitly by the user or by a teacher of the system, through dedicated inputs, or be learned by the system, through the input discourse. It should be the same for place and time information. These author or hypotheses information will also be used to build models of the various partners of the system. These models will propose an assumed behavior of these partners, in the context of one or several concepts. A concept is a class of situations or a set of sentences or a class of words.

A specific author represents the system. Its model enables it to make inferences on its own behavior or concepts. An assumption may of course have an author (or a class of authors), which may be the system itself.

We will notice the importance of images of the self and of others, and of emotions, in the learning of the outside world by an intelligent system. These images (or models) can make possible an imprinting mechanism, by which the system will adopt the values and concepts of its designer, or of a teacher who plays a privileged role in its development. The adoption of the values of a predefined group, may also play a role in this sense.

We may consider that if there had been an imprinting mechanism or an image of third parties in Tay, the intelligent system experimented by Microsoft on Twitter, this system would have identified the discourse of hostile third parties or of people with values opposed to those of its group or designers, and would have limited it to these third parties [33].

## 4.8. Spatial representation

Ideally, the system should be able to spot the presence of an agent or any object on one or several maps, at various scales, and recognize which objects are situated in various neighborhoods of itself. Map management operators should then be available. These operators could thus manage interfaces with external memory, which would hold these maps.

However in a first step it seems possible to manage this issue through the representation of inclusion relationships of an object in another one, as the system will have a known capability to process trees.

## 4.9. Time representation

Many different representations of time may be present in texts, such as relative ones "before this event", "the day before", or imaginary times "once upon a time", or different time scales "one millisecond after the Big Bang".

The time may be represented by a scale of time, and by a number, which represents the distance to an origin of time, and by a duration. It is also necessary to take into account two distinct times : the event time, and the time of its recording or description. E.g., an event may relate to the origin of the Universe, more than 13 billion years ago, and may be described in an article published in "December 2015".

However, one issue is the representation of this notion, in a RNN architecture. For the moment, we don't have fully resolved this point. A provisional solution is similar to that which we have proposed for the space representation: it is to represent a tree of time intervals, with operators of inclusion, precedence and succession.

## 4.10.     Degree of confidence in a data or a property.

A very important information in an AGI is the degree of confidence that the system may have towards a sentence, or a model, an assumption, a logical clause, an author, etc. This confidence level may itself be decomposed into several elements, such as the evaluation of the sentence, the level of confidence of this evaluation, the cardinality of the validation set (the number of examples), etc.

As an example, some opponent I of the system will be characterized with certainty C, to produce declarations with value V, with an experimental basis X (number of examples). C describes the degree of knowledge that the designer or observer O has about I, the level of X indicates the plausibility of further revisions of these data. V describes the community or opposition of values between I and O.

In the same way, an assumption H can be characterized by the triple (C, X, V). If these variables are defined, without loss of generality, between 0 and 1, the author of a new hypothesis will possibly characterize it by the triple {0.9, 0.1, 0.7}. Her confidence in the assumption will be C= 0.9, the experimental basis will possibly be X= 0.1, and the agreement between the author and O may be V= 0.7. It will be possible to make some calculations on these coefficients, the calculations will be more or less approximate.

In the neuronal system which we describe here, the confidence coefficient may be reduced to a scalar, which is the arc value between the input character (INPUT) and the entry layer of the RNN. The possibility to use a multi-valued confidence coefficient should be the subject of further research.

## 4.11.     Non-monotony, Open World Assumption (OWA)

An important property of a reflexive system is non-monotony, which enables the calculation of clauses with negation. This property enables exceptions to the knowledge acquired by the system, or also systems of concepts (ontologies) with only partial coherence.

This property also makes possible a progress in the system knowledge, by enabling detection of inconsistencies in a theory..

In a neuron network, the non-monotony can be implemented by the use of inhibitor arcs between two neurons, i.e. the use of negative connection weights (synaptic weights). This implies an adaptation of the retro-propagation algorithm, or more generally the learning algorithm.

The non-monotony also implies the Open World Assumption (OWA), i.e. the assumption that the absence of a clause in the knowledge base (or of a connection in the neural network) does not imply its negation.

## 4.12.     Multimedia perception

**Definition 4.**
*We shall say that a system is fully reflexive if it includes multimedia perceptions, with an input of documents (text, images, page layout), visual data (still image and video), audio data (sounds and speech), and is able to process them.*

In the case of an embodied system (robot or autonomous vehicle) it will possibly include other parameters, such as the gps position and localization on a map.

## 4.13.     Actions and physical support

**Definition 5.**
*We shall say that an autonomous system is complete if it is a reflexive and embodied system [12], able of mobility and with a set of actions implemented by effectors (limbs, mobility systems), under the control of the reflexive system.*

However the system will be more or less complete, depending on the properties of its mobility system and of its effectors.

## 4.14.     External interfaces

The access of a neuron network, possibly recurrent, to external memory (and more generally to external interfaces) is the subject of recent work. These interfaces may enable an access to sets of registers, stacks, or random access memory [18, 20, 42]. The access can also be implemented in an associative way, by one or several keywords produced by the system.

We assume that the external documents accessed by the system can be memorized in an external memory, and that the system can access it by one of these methods to answer a user query.

## 5.   ARCHITECTURE

The architecture of the intelligent system which we consider here, will be characterized by a first level of n memorization cells, making a 1D systolic network, in which the user will input a character at a time. This network will possibly be connected to a stack of N registers of n characters each. The operations on this stack will be a subset of the actions of the system.

The n cells of the first layer will be connected to jmax layers of h(j) neurons each, forming a recurrent network. The output of layer jmax will include first, the k actions of the system, and secondly, an output of c characters. Both elements will form the retroaction loop of the system, and will be connected to its entry layer, beside the sentence contained in the first n memorization cells.

The coupling of this unit with another equivalent recurrent unit, according to the schema of paragraph 3, is intended to enable a degree of deliberation. It will be the subject of further experimentation.

## 6.   INITIALIZATIONS

The initialization will be made by the choice of a global architecture among the few variants presented before, and of the meta-parameters such as the number of neurons per layer, the number of neurons per layer and cognitive area, the projections between cognitive areas, the initial synaptic weights. The functions which will return the bias depending on the level of the various emotions, will also be part of initializations.

We shall assume that when the list of connections will be defined, the initial weights will be drawn according to a simple law, which remains to be defined.

It is possible to evolve these meta-parameters, through the use of an evolutionary program. However, this leads to two issues:

First, the choice of a success criterium of a set of parameters, such a criterium is needed to guide the evolution program. It can be the number of iterations needed to reach some quality of results, as measured by subjective indices. These results may also be used to measure the quality of the reflexivity, as we propose it in paragraph 8.

Secondly, Eliezer Yudkowsky stresses the risk, if an AGI is defined by an evolutionary program, that it would escape to the control of its designers [38, 39]. This risk should be carefully contained.

A solution may be in a first time, to reserve these evolutionary programs to system versions containing relatively few neurons (e.g. less than 100 k, or possibly even less than 10 k), and therefore for which the risk of escape would appear neglectible. This remains to be worked out.

As an example, a system with 30 to 50 k neurons has a comparable size to the brain (or equivalent neural system) of an  insect with a limited complexity, such as the fly. Even if the functions of neurons in an AGI of comparable cardinality are very different from those of a fly, the considered risk would probably remain very low. It might not be the case for cardinalities about or above one million neurons, comparable to very evolved insects like the honey bee, which can recognize faces and have complex social behavior [14, 28].

## 7.   EXPERIMENTATION

An implementation of the system in Python is scheduled for the next months, it will use Theano libraries. It should be first implemented on an OS X environment, with relatively low processing power, then in a CUDA or OpenCL environment on a Mac Pro, with a D700 graphic board (more than 4000 GPU cores).

However the possibility of implementing part of the initial development and experimentation on Ubuntu, is also considered as an alternative solution.

The goal of experimentation will be to work out a system which should be able to learn from Wikipedia and text bases available on the Internet, and to participate to a process of Queries & Answers. The system should learn the concepts and values associated with several authors, and will be able to take them into account to evaluate queries and answers. We believe that this accounting of the model of discussion partners is essential for a mastered learning of the outside world.

## 8.   REFLEXIVITY EVALUATION.

We believe that the proposed architecture is a minimal one to obtain a perceivable degree of reflexivity. Let us remember that our goal is not to realize a "conscious" system, but to simulate some properties of the cognitive systems of human beings and of some superior animals (primates, mammals, some birds…), or maybe of some simpler animals (drosophila, honeybee) [10, 14, 28].

We propose the following conjectures:

**Conjecture 1.**

*A necessary condition for a neuron network to implement a reflexive system is that it should possess the properties described in paragraph 2.*

**Conjecture 2.**

*A sufficient condition for an AGI to be a complete reflexive autonomous system is that it would possess all the properties described in paragraphs 2, 3, and 4.1 to 4.14.*

We shall distinguish complete reflexive autonomous systems (r-complete systems), and complete deliberative autonomous systems (d-complete systems). A **conjecture 2-b** applies to d-complete systems.

Let's notice that an AGI can be a complete reflexive system without being an autonomous one. We should also stress the fact that, in our opinion, the condition of conjecture 2 is sufficient, but not necessary. We will have to refine the analysis and the experimentation in order to reach necessary and sufficient conditions.

We will provide a reflexivity index, which will evaluate the satisfaction of conditions from paragraphs 2 to 4.14, in order to determine to which degree an intelligent system is reflexive. The exact composition of the reflexivity index will be the subject of further papers.

**Conjecture 3.**
*The properties of a complete reflexive AGI, or a complete autonomous one, can be obtained either with a neuronal system, or with a partially or fully symbolic system.*

Let us notice that it seems easier to implement a system according to the conditions of paragraphs 2 to 4.14 with a neuronal system (e.g. to implement non-monotony, fuzzy inference, etc.), but that the interest of a symbolic system is that it could be understandable by users, which is not necessarily the case with neuronal systems.

An index of the properties of an intelligent system may aggregate coefficients describing each dimension of an AGI, such as presented in paragraphs 2 to 4.14. E.g. each coefficient may be comprised between 0 and 1, and the global coefficient may be a product of these partial coefficients, or the sum of their logarithms, plus some constant. The basic properties of reflexivity and deliberation may however have larger weights.

A second complementary index may be the cardinality of the neurons set of the upper layer (or the upper layer of the encoder part), or some function of this cardinality.

It will be possible to validate this approach by comparing these coefficients with the result of subjective evaluations [23], e.g. a Turing test, applied to a relatively dense system (in terms of number of neurons and of connections per neuron), after a learning phase, which may be based on a subset of the Wikipedia pages.

## 9. CONCLUSION

In this paper, we have proposed several conjectures on the properties that should be those of a neuron network, or possibly of a symbolic architecture, to implement a reflexive and deliberative system, and possibly a reflexive and autonomous system.

In the next months we intend to experiment a neuronal architecture which will display these properties. The neuronal network which will be implemented will not exceed the cardinality of the neural system of a drosophila, or possibly of a honeybee, i.e. cardinalities between 30 k and one million of neurons, and a number of weights between 30 million and 10 billion, or more probably not greater than one billion weights.

However, with an appropriate topology, this system should be applicable to problems which interest people, such as the summarization of Wikipedia pages or the participation to a query and answers system.

## ACKNOWLEDGEMENTS

The author thanks Jean-François Perrot and Francis Wolinski for useful comments.

## REFERENCES

1. S.S. Adams, I. Arel, J. Bach, R. Coop, R. Furlan, B. Goertzel, et al., Mapping the Landscape of Human-Level Artificial General Intelligence, *AI Magazine*, Spring 2012, 25-41

2. B. Baar, *A cognitive theory of consciousness*, Cambridge Univ. Press, July 30, 1993

3. J. Bach, *Principles of Synthetic Intelligence*, Oxford Univ. Press, 2009

4. J. Bach, Modeling Motivation in MicroPsi2, in *Artificial General Intelligence,* Proceedings 8th International Conference, Berlin, Germany, July 2015, LNAI 9205, Springer 2015

5. J. Barrat, *Our Final Invention: Artificial Intelligence and the End of the Human Era*, Thomas Dunne Books, Saint Martin Press, NY, USA 2013

6. J. Bieder, B. Goertzel, A. Potapov (editors), *Artificial General Intelligence*, Proceedings 8[th] Intl. Conf., AGI 2015, Berlin, Germany, July 22-25, 2015, LNAI 9205, Springer, 2015

7. N. Boström, *SuperIntelligence, Paths, Dangers, Strategies*, Oxford University Press, UK, 2014

8. D.J. Chalmers, Facing Up to the Problem of Consciousness, *Journal of Consciousness Studies*, **2**(3): 200-19, 1995

9. S. Dehaene, *Consciousness and the Brain: Deciphering How the Brain Codes our Thoughts*, Penguin Books, Jan. 2014

10. W.T. Gibson et al., Behavioral Responses to a Repetitive Visual Threat Stimulus Express a Persistent State of Defensive Arousal in Drosophila, *Current Biology*, 2015

11. B. Goertzel, Artificial General Intelligence: Concept, State of the Art, and Future Prospects, *Journal of Artificial General Intelligence*, **5**(1) 1-46, 2014

12. B. Goertzel, C. Pennachin, N. Geisweiller, *Engineering General Intelligence, Part2, The CogPrime Architecture for Integrative, Embodied AGI*, Atlantis Press, Paris, France, 2014

13. B. Goertzel, L. Orseau, J. Snaider (editors), *Artificial General Intelligence*, Proceedings 7th Intl. Conf., AGI 2014, Quebec City, QC, Canada, August 1-4, 2014, LNAI 8598, Springer 2014

14. J.-L. Goudet, *Reconnaissance des visages : l'ordinateur devrait imiter... les abeilles* (in French) 2-2-2009, in www.futura-sciences.com , read on 20-5-2015

15. S. Hawking, et al., Transcendence looks at the implications of artificial intelligence. But are we taking AI seriously enough ? *The Independent,* 20 May 2015

16. H.P. Helgason, K.R. Thorisson, D. Garrett, E. Nivel, "Towards a General Attention Mechanism for Embedded Intelligent Systems", *Intl. J. of Computer Science and Artificial Intelligence*, Mars 2014, **4**, 1, 1-7

17. J. Hernández Orallo, « Evaluation of Intelligent Systems », Tutorial, *AGI-2015*, Berlin, Germany, July 2015

18. A. Joulin, T. Mikolov, Inferring Algorithmic Patterns with Stack-Augmented Recurrent Nets, Facebook AI Research, arXiv:1503.01007v4 [cs.NE] 1 Jun 2015

19. A. Karpathy, The Unreasonable Effectiveness of Recurrent Neural Networks, in *Andrej Karpathy blog,* May21, 2015, http://karpathy.github.lo/2015/05/21/rnn-effectiveness/

20. K. Kurach, M. Andrychowicz, I. Sutskever, Neural Random-Access Machines, Google, arXiv1511.06392v2 [cs.LG] 7 Jan 2016

21. Y. Le Cun, L'apprentissage profond, course at Collège de France, chaire Informatique et Sciences Numériques, 2015-2016, http://www.college-de-france.fr (in French)

22. H. Lövheim, A new three-dimensional model for emotions and monoamine neuro-transmitters, *Med. Hypotheses*, 2012, **78**, 341-348

23. G. Marcus, F. Rossi, M. Veloso, (editors) Beyond the Turing Test, special issue of *AI Magazine*, Spring 2016

24. J. McCarthy, Making Robots Conscious of their Mental States, Stanford University, July 24, 1995, http://www-formal.stanford.edu/jmc/ published in *Machine Intelligence* 15, Muggleton S. (Ed), Oxford Univ. Press, 1996

25. J. McCarthy, Awareness and Understanding in Computer Programs, a Review of "Shadows of the Mind", by Roger Penrose, *Psyche*, **2**(11), July 1995

26. M. Minsky, *The Emotion Machine* (Commonsense thinking, artificial intelligence and the future of the human mind), Simon and Schuster Ed., NY, USA 2006

27. V. Mnih, N. Hees, A. Graves, K. Kavukcuoglu, Recurrent Models of Visual Attention, Google DeepMind, arXiv:1406.6247v1 [cs.LG] 24 Jun 2014

28. J.-Y. Nau, Les abeilles, bonnes physionomistes. *Slate*, 4-2-2010 (in French)

29. R. Penrose, *The Emperor's New Mind: Concerning Computers, Minds, and The Laws of Physics,,* Oxford Univ. Press, Oxford, UK, 1989

30. R. Reddy, Creating Human Level AI by Educating a Child Machine, in The Convergence of Machine and Biological Intelligence, *IEEE Intelligent Systems*, 2013

31. M.O. Riedl, B. Harrison, Using Stories to Teach Human Values to Artificial Agents, Georgia Institute of Technology, AAAI 2015

32. J. Schmidhuber, Deep Learning In Neural Networks, an Overview, *Neural Networks* **61**, 85-117 (2015, Online 2014)

33. M. Tual, A peine lancée, une intelligence artificielle de Microsoft dérape sur Twitter, Le Monde.fr | 24.03.2016 (in French)

34. O. Vinyls, L. Kaiser, T. Koo, S. Petrov, I. Sutskever, G. Hinton, Grammar as a Foreign Language, Google, arXiv:1412.7449v3 [cs.CL] 9 Jun 2015

35. P. Wang, Motivation Management in AGI Systems, Temple University, Philadelphia, USA, *Proceedings of AGI-12*, Oxford, UK, December 2012

36. P. Wang, Analogy in a General Purpose Reasoning System, *Cognitive Systems Research*, **10**-(3), 286-296, 2009

37. Wikipedia, « neurotransmetteurs » (ou neuromédiateurs), read on 7-4-2016 (In French)

38. E. Yudkowzky, Artificial Intelligence as a Positive and Negative Factor in Global Risk, in *Global Catastrophic Risks*, edited by Nick Boström and Milan M Ćircović, 308-345, Oxford UP, NY 2008

39. E. Yudkowsky, Creating Friendly AI 1.0 : The Analysis and Design of Benevolent Goal Architectures, The Singularity Institute, San Francisco, USA, June 2001

40. E. Yudkowsky, General Intelligence and Seed AI-Creating Complete Minds Capable of Open-Ended Self-Improvement, Research Report, MIRI, 2001

41. E. Yudkowsky, Complex Value Systems In Friendly AI, *Artificial General Intelligence* 4th International

Conference, AGI 2011, USA Aug 3-6 2011
Proceedings, in *LNCS* vol **6830**, Springer, Berlin 2011

42.  W. Zaremba, I. Sutskever, Reinforcement Learning
Neural Turing Machines (revised), Facebook AI
Research & Google Brain, *submitted to ICLR 2016*,